\documentclass[conference]{IEEEtran}

\usepackage{algorithm}
\usepackage{algpseudocode}
\usepackage{amsmath}
\usepackage{amssymb}
\usepackage{graphicx}
\usepackage{lettrine}
\usepackage{multirow}
\usepackage{threeparttable}

\begin{document}

\title{An Inter-molecular Adaptive Collision Scheme for Chemical Reaction Optimization}

\author{James J.Q. Yu,
       \textit{Student Member, IEEE} and
       Victor O.K. Li,
       \textit{Fellow, IEEE}\\
       Department of Electrical and Electronic Engineering\\
       The University of Hong Kong\\
       Email: \{jqyu, vli\}@eee.hku.hk\\
\and Albert Y.S. Lam,
       \textit{Member, IEEE}\\
       Department of Computer Science\\
       Hong Kong Baptist University\\
       Email: albertlam@ieee.org}

\maketitle
\pagestyle{empty}

\begin{abstract}
Optimization techniques are frequently applied in science and engineering research and development. Evolutionary algorithms, as a kind of general-purpose metaheuristic, have been shown to be very effective in solving a wide range of optimization problems. A recently proposed chemical-reaction-inspired metaheuristic, Chemical Reaction Optimization (CRO), has been applied to solve many global optimization problems. However, the functionality of the inter-molecular ineffective collision operator in the canonical CRO design overlaps that of the on-wall ineffective collision operator, which can potential impair the overall performance. In this paper we propose a new inter-molecular ineffective collision operator for CRO for global optimization. To fully utilize our newly proposed operator, we also design a scheme to adapt the algorithm to optimization problems with different search space characteristics. We analyze the performance of our proposed algorithm with a number of widely used benchmark functions. The simulation results indicate that the new algorithm has superior performance over the canonical CRO.
\end{abstract}


\section{Introduction}
\lettrine[lines=2]{C}{hemical} Reaction Optimization (CRO) \cite{LamLi2010ChemicalReactionInspired} is a recently proposed population-based general-purpose metaheuristic optimization method. CRO mimics the interactions and transformations of the reactant molecules in chemical reactions to search for global optimums \cite{LamLi2012ChemicalReactionOptimization:}. CRO was designed as a general framework for optimization, and it was initially proposed to solve combinatorial optimization problems \cite{LamLi2010ChemicalReactionInspired}. It has been applied to solve a wide range of classical and real-world discrete problems, e.g. Quadratic Assignment Problem \cite{LamLi2010ChemicalReactionInspired}, Sensor Deployment Problem \cite{YuLiLam2012SensorDeploymentAir}, and Unit Commitment Problem \cite{YuLiLam2013OptimalV2GScheduling}. CRO demonstrated outstanding performance in these problems and was shown to be both effective and efficient in solving similar combinatorial problems.

Based on the CRO optimization framework, Lam \textit{et al.} proposed a variant of CRO, named Real-Coded Chemical Reaction Optimization (RCCRO), to solve continuous optimization problems \cite{LamLiYu2012RealCodedChemical}. RCCRO has four different operators for the elementary reactions to facilitate the optimization process (the details will be introduced in Section \ref{sec:CRO}). The encoding scheme, boundary handling scheme, and other ancillary schemes were also devised. RCCRO has also been applied to solve different benchmark and real-world problems, e.g. training artificial neural networks \cite{YuLamLi2011EvolutionaryArtificialNeural}, Optimal Power Flow problem \cite{SunLamLiXuYu2012ChemicalReactionOptimization}, and Cognitive Spectrum Allocation problem \cite{LamLiYu2013PowerControlledCognitive}. We also conducted research on exploring the nature of different operators in RCCRO. In our previous work, we attempted to use different probability distribution function to replace the Normal distribution adopted to perform neighborhood search \cite{YuLamLi2012RealcodedChemical}, and the results show no significant preference over the four distributions studied.

In the canonical design of RCCRO, the inter-molecular ineffective collision operator is largely similar to having two on-wall ineffective collisions occurring simultaneously, and the only difference between inter-molecular and on-wall ineffective collisions is their different energy handling schemes. This design can significantly alleviate the implementation effort as the neighborhood search operator, i.e. on-wall ineffective collision operator, is re-used. However, a potential drawback of this implementation is that the functionalities of these two elementary reactions overlap, which may lead to wasting the limited function evaluations. Moreover, the neighborhood search operator in the canonical RCCRO alters the value of one dimension of the input solution, which can effectively solve problems with minimal inter-dimensional correlation. But it is not efficient at solving problems with strong inter-dimensional correlation such as rotated functions. In order to resolve these drawbacks, this paper focuses on proposing new operators and schemes based on the canonical RCCRO framework.

The rest of this paper is organized as follows. We introduce the canonical design of the CRO framework and the RCCRO design in Section \ref{sec:CRO}. The newly proposed inter-molecular ineffective collision operator and the adaptive collision scheme is presented in Section \ref{sec:CROAC}. Section \ref{sec:expsim} introduces the experimental setting we employed to analyze the performance of our algorithm as well as the simulation results and discussions. Finally this paper is concluded in Section \ref{sec:conclusion} with some potential future research.

\section{Chemical Reaction Optimization Framework} \label{sec:CRO}

In this section we will introduce the general framework of the canonical CRO and some implementation details of RCCRO. We will use CRO to refer to both CRO and RCCRO hereafter.

\subsection{Molecule}

Molecules are the basic operating agents of CRO. In CRO, each optimization task is considered as a chemical reaction occurring in a closed container with a population of molecules and an energy buffer. The molecules in the container move and collide with the wall or with other molecules. With the collisions, the properties of molecules involved are changed, allowing the algorithm to explore the solution space. Each molecule possesses the following attributes:

\begin{itemize}
\item A molecular structure refers to a feasible solution of the problem to be optimized. Each molecule possesses one molecular structure, which is modified by different operators invoked in the elementary reactions.
\item Potential Energy (PE) stands for the quality of the solution held by the molecule. In terms of optimization, PE is equivalent to the fitness value of the solution.
\item Kinetic Energy (KE) stands for the tolerance of the molecule to accept a worse molecular structure compared to the current one in terms of the solution quality. A larger KE means that the molecule can accept a much worse molecular structure than those molecules with less KE values.
\end{itemize}

Besides these properties, the molecules can also hold some other attributes to suit different optimization problems. The implementation of these attributes can be problem-independent.

\subsection{Elementary Reactions}

In CRO, the optimization task, i.e. the chemical reaction, is tackled by four kinds of elementary reactions, including on-wall ineffective collision (\textit{Onwall}), decomposition (\textit{Dec}), inter-molecular ineffective collision (\textit{Inter}), and synthesis (\textit{Syn}). They occur sequentially and randomly to manipulate the structures of the molecules involved. Among them, \textit{Onwall} and \textit{Dec} takes one molecule as input while \textit{Inter} and \textit{Syn} takes two molecules. \textit{Onwall} and \textit{Syn} output one molecule based on the input molecule(s), while \textit{Dec} and \textit{Inter} output two molecules. Thus \textit{Dec} and \textit{Syn} can alter the total population size. This dynamic population size property is a feature that distinguishes CRO from other metaheuristics.

\subsubsection{Neighborhood Search Operator}

In the canonical design of CRO, we use this neighborhood search operator as the basic molecular structure manipulatimg operator \cite{LamLi2012ChemicalReactionOptimization:}. It is employed in \textit{Onwall}, \textit{Dec}, and \textit{Inter}. This operator modifies one molecular structure using Gaussian perturbation value. Assume that the original molecular structure is $\omega$, and the newly generated molecular structure is $\omega^\prime$. We first randomly choose a dimension $i$ of $\omega$ to modify. Then a random number $\epsilon$ is generated from a zero-mean Normal distribution whose variance is a user-defined parameter $\textit{stepSize}$. Then the new molecular structure is generated by
\begin{equation}
\omega^\prime_i = \omega_i + \epsilon
\end{equation}
where $\omega_i$ stands for the $i$-th dimension of the molecular structure $\omega$.

\subsubsection{Elementary Reaction Operators}

Based on the neighborhood search operator, we further designed the four elementary reaction operators. The detailed implementations are elaborated as follows.

\begin{itemize}
\item An \textit{Onwall} occurs when a molecule collides with the container and bounce back. During this elementary reaction, the structure of the involved molecule is manipulated using the neighborhood operator. The main purpose of this elementary reaction is to perform local search.

\item A \textit{Dec} occurs when a molecule collides with the container and breaks into two new molecules. During this elementary reaction, the structure of the input molecule is copied to the two new molecules. Then the new molecules go through a number of independent neighborhood operators. The main purpose of this elementary reaction is to jump out of local optimum.

\item An \textit{Inter} occurs when two molecules collide with each other and bounce back. In the canonical implementation of CRO, this elementary reaction is considered to be two \textit{onwalls} occurring simultaneously. The main purpose of this elementary reaction is also to perform local search.

\item A \textit{Syn} occurs when two molecules collide and merge into one new molecule. The detailed implementation of this elementary reaction can be found in \cite{LamLi2012ChemicalReactionOptimization:}. The main purpose of this elementary reaction is to maintain the population diversity \cite{YuLamLi2012RealcodedChemical}.
\end{itemize}

\subsection{Searching Pattern}

CRO is composed of three phases: initialization, iteration, and the final phase. In initialization, all algorithm parameters are initialized, and the initial population is randomly generated. Then in the iteration phase, the algorithm manipulates the molecules to search for the global optimum in an iterative manner. In each iteration one elementary reaction is selected and conducted. This selection is controlled by the algorithm optimization parameters using a pre-defined deterministic scheme \cite{LamLi2012ChemicalReactionOptimization:}. After the elementary reaction is selected, a corresponding number of molecules are randomly picked to participate in the reaction. These molecules are then manipulated and go through an energy check to determine whether the reaction succeeds or not, i.e. whether the changes are adopted. The energy check is elaborated in \cite{LamLi2010ChemicalReactionInspired} and \cite{LamLiYu2012RealCodedChemical}, and the underlying concept is the conservation of energy. This phase iterates until the stopping criterion is met. Then the algorithm proceeds to the final phase and the optimum results are output. Interested readers can refer to \cite{LamLi2010ChemicalReactionInspired}, \cite{LamLi2012ChemicalReactionOptimization:}, and \cite{LamLiYu2012RealCodedChemical} for elaboration of the algorithm and its pseudocode.

\section{Chemical Reaction Optimization with Adaptive Collision} \label{sec:CROAC}

In this section we will elaborate our proposed inter-molecular adaptive collision scheme, which is composed of a new inter-molecular ineffective operator and an adaptive collision scheme. We call this algorithm CRO with Adaptive Collision (CRO/AC). In all discussions hereafter, without loss of generality, we only consider minimization problems.

\subsection{Inter-molecular Ineffective Collision Operator}

In the canonical design of CRO, the inter-molecular ineffective collision is considered to be two on-wall ineffective collisions occurring simultaneously. However, this implementation renders the effects of these two operators overlapping. So we propose a new \textit{Inter} operator to overcome this drawback.

When an \textit{Inter} takes place, two molecules are randomly selected from the current population to participate in the elementary reaction. Instead of conducting neighborhood operators on them as in canonical CRO, we first check their PE values, i.e. the fitness values of the solutions they hold. We use $\omega_s$ to denote the molecular structure of the molecule with a larger PE value, and $\omega_t$ to denote the one with a smaller PE value. This implies that $\omega_t$ is better than $\omega_s$ in terms of the solution quality.

After the two molecules are examined, we use a two-step approach to modify their structures. In the first step, we change $\omega_s$ to make it more similar to $\omega_t$. This can be accomplished by using the following:
\begin{equation}
\omega^\prime_{s,i} = (\omega_{t,i} - \omega_{s,i}) \times r_i + \omega_{s,i},
\end{equation}
where $\omega_{s,i}$ stands for the value of $i$-th dimension of $\omega_s$, $\omega^\prime_{s,i}$ stands for the newly generated structure for $\omega_s$, and $r_i$ is a random number uniformly generated in $[0,1)$. In this step, the random number $r_i$ is independently generated for each dimension.

The second step is to change $\omega_t$ to make it less similar to $\omega^\prime_s$, accomplished by:
\begin{equation} \label{eqn:interstep2}
\omega^\prime_{t,i} = (\omega_{t,i} - \omega^\prime_{s,i}) \times r_i + \omega_{t,i}.
\end{equation}
The purpose of this design is to avoid pre-mature convergence and maintain the population density. The first step significantly reduces the population density, which may lead to pre-mature convergence. This second step increases the density by a small margin, and acts as a counter-measure of getting stuck in local optimums. We will demonstrate the necessity of this step in Section \ref{sec:expsim}.

The remaining parts of the \textit{Inter} operator are unchanged. The new molecules will go through the energy conservation test to check whether this \textit{Inter} is deemed successful or not. If so, the changes made on these molecules, including molecular structures and molecule properties, are accepted. Otherwise these molecules remain unchanged. This ends an \textit{Inter} elementary reaction for CRO.

\subsection{Adaptive Collision Scheme}

One main difference of our proposed \textit{Inter} operator and the canonical design is that our operator can update the values of multiple dimensions in a solution. This operator is suitable for problems with large inter-dimensional correlations \cite{QinHuangSuganthan2009DifferentialEvolutionAlgorithm}, but may have equal or even poorer performance on those problems with no such correlations. Thus the ratio of occurrence of \textit{Onwall} and \textit{Inter} is critical to the performance. In the canonical design of CRO, this ratio is generally controlled by a user-defined parameter collision rate $collRate$ \cite{LamLi2010ChemicalReactionInspired}, and the value is usually set to be 0.2 \cite{LamLiYu2012RealCodedChemical}\cite{XuLamLi2011ChemicalReactionOptimization}. However, with the changes in the nature of \textit{Inter} operator, this value is no longer appropriate and we shall tune the parameter to tailor-make the algorithm  for different optimization problems, but this tuning process can be very time-consuming. So we further devise an adaptive collision scheme for CRO to control the ratio of collision adaptively, using the information feedback from the optimization process \cite{BeyerDeb2001SelfAdaptiveFeatures}.

\begin{figure}
    \includegraphics[width=0.48\textwidth]{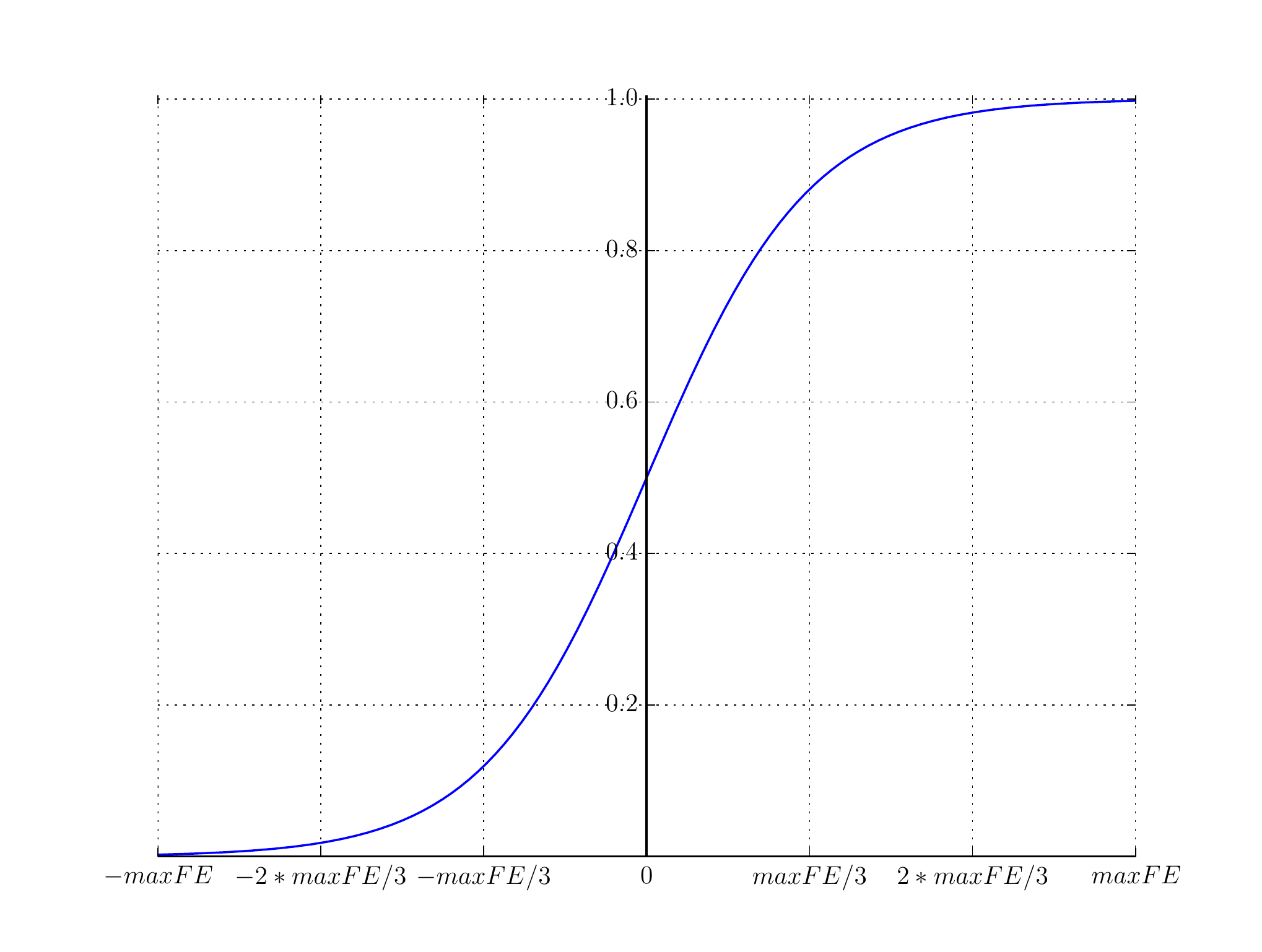}
    \caption{Sigmoid Function Used in Adaptive Collision Scheme}
    \label{fig:sigmoid}
\end{figure}

In our adaptive collision scheme, we add a new attribute to the system, namely the number of successful \textit{inters}, denoted $counter$. Initially, this attribute is set to zero, and at the beginning of each iteration, the value of $collRate$ is calculated using the following sigmoid function:
\begin{equation}
collRate = \frac{1}{1 + \exp(-6\times\frac{counter}{maxFE})},
\end{equation}
where $maxFE$ is the maximum allowed number of function evaluations. The plot of this function is demonstrated in Fig. \ref{fig:sigmoid}, where the x-axis is the value of $counter$ and the y-axis is the value of the corresponding $collRate$. 
The value of $counter$ is updated whenever an \textit{Onwall} or \textit{Inter} occurs and succeeds. If a successful \textit{Onwall} occurs, the value of $counter$ is reduced by one. Otherwise the value is increased by one. In this way, the algorithm can adjust the $collRate$ to favor either \textit{Onwall} or \textit{Inter}, to match the optimization problem being solved.

\begin{table*}
	\caption{Benchmark Functions}
	\label{tbl:benchmark}
	\begin{center}
        \scriptsize
        \begin{threeparttable}
            \begin{tabular}{llll}
                \hline
                $f$ & Function & Solution Transformation\tnote{*} & Name \\
                \hline
                $f_1$ & $\begin{aligned}\sum\nolimits^D_{i=1}z^2_i\end{aligned}$ & $\mathbf{z} = \mathbf{x} - \mathbf{o}$ & Shifted Sphere Function \\
                $f_2$ & $\begin{aligned}\sum\nolimits^D_{i=1}\sum\nolimits^i_{j=1}z^2_i\end{aligned}$ & $\mathbf{z} = \mathbf{x} - \mathbf{o}$ & Shifted Schwefel's Problem 1.2 \\
                $f_3$ & $\begin{aligned}\sum\nolimits^D_{i=1}|z_i|+\prod\nolimits^D_{i=1}|z_i|\end{aligned}$ & $\mathbf{z} = (\mathbf{x} - \mathbf{o}) \times 0.1$ & Shifted Schwefel's Problem 2.22 \\
                $f_4$ & $\begin{aligned}\sum\nolimits^D_{i=1}|z_i|+\prod\nolimits^D_{i=1}|z_i|\end{aligned}$ & $\mathbf{z} = \mathbf{M}(\mathbf{x} - \mathbf{o}) \times 0.1$ & Shifted Rotated Schwefel's Problem 2.22 \\
                $f_5$ & $\begin{aligned}-20\exp(-0.2\sqrt{\frac{1}{n}\sum\nolimits^n_{i=1}z^2_i}) - \exp(\frac{1}{n}\sum\nolimits^n_{i=1}\cos 2\pi z_i)\\ + 20 + e\end{aligned}$ & $\mathbf{z} = (\mathbf{x} - \mathbf{o}) \times 0.32$ & Shifted Ackley's Function \\
                $f_6$ & $\begin{aligned}-20\exp(-0.2\sqrt{\frac{1}{n}\sum\nolimits^n_{i=1}z^2_i}) - \exp(\frac{1}{n}\sum\nolimits^n_{i=1}\cos 2\pi z_i)\\ + 20 + e\end{aligned}$ & $\mathbf{z} = \mathbf{M}(\mathbf{x} - \mathbf{o}) \times 0.32$ & Shifted Rotated Ackley's Function \\
                $f_7$ & $\begin{aligned}418.9829D-\sum\nolimits^D_{i=1}z_i\sin\sqrt{|z_i|}\end{aligned}$ & $\mathbf{z} = \mathbf{x} \times 5$ & Schwefel's Problem 2.26\\
                $f_8$ & $\begin{aligned}418.9829D-\sum\nolimits^D_{i=1}z_i\sin\sqrt{|z_i|}\end{aligned}$ & $\mathbf{z} = \mathbf{M}\mathbf{x} \times 5$ & Rotated Schwefel's Problem 2.26\\
                $f_9$ & $\begin{aligned}\sum\nolimits^D_{i=1}(z^2_i - 10\cos 2\pi z_i + 10)\end{aligned}$ & $\mathbf{z} = (\mathbf{x} - \mathbf{o}) \times 0.0512$ & Shifted Rastrigin's Function\\
                $f_{10}$ & $\begin{aligned}\sum\nolimits^D_{i=1}(z^2_i - 10\cos 2\pi z_i + 10)\end{aligned}$ & $\mathbf{z} = \mathbf{M}(\mathbf{x} - \mathbf{o}) \times 0.0512$ & Shifted Rotated Rastrigin's Function\\
                $f_{11}$ & $\begin{aligned}\sum\nolimits^D_{i=1}\frac{z^2_i}{4000}-\prod^D_{i=1}\cos\frac{z_i}{i}+1 \end{aligned}$ & $\mathbf{z} = (\mathbf{x} - \mathbf{o}) \times 6$ & Shifted Griewank's Function\\
                $f_{12}$ & $\begin{aligned}\sum\nolimits^D_{i=1}\frac{z^2_i}{4000}-\prod^D_{i=1}\cos\frac{z_i}{i}+1 \end{aligned}$ & $\mathbf{z} = \mathbf{M}(\mathbf{x} - \mathbf{o}) \times 6$ & Shifted Rotated Griewank's Function\\
                $f_{13}$ & $\begin{aligned}\sin^2(\pi y_1) + \sum\nolimits_{i=1}^{n-1}[(y_i-1)^2(1+10(\sin^2y_{i+1}))] + \\(y_n-1)^2(1+\sin^2(2\pi y_n)), y_i = 1+\frac{1}{4}(z_i+1) \end{aligned}$ & $\mathbf{z} = (\mathbf{x} - \mathbf{o}) \times 0.1$ & Shifted Levy's Function\\
                $f_{14}$ & $\begin{aligned}\sin^2(\pi y_1) + \sum\nolimits_{i=1}^{n-1}[(y_i-1)^2(1+10(\sin^2y_{i+1}))] + \\(y_n-1)^2(1+\sin^2(2\pi y_n)), y_i = 1+\frac{1}{4}(z_i+1) \end{aligned}$ & $\mathbf{z} = \mathbf{M}(\mathbf{x} - \mathbf{o}) \times 0.1$ & Shifted Rotated Levy's Function\\
                $f_{15}$\tnote{+} & $\begin{aligned}\frac{1}{10}[\sin^2(3\pi z_1)+\sum\nolimits_{i=1}^{n-1}(z_i-1)^2(1+\sin^2(3\pi z_{i+1}))+\\(z_n-1)^2(1+\sin^2(2\pi z_n))]+\sum\nolimits_{i=1}^n u(z_i,5,100,4) \end{aligned}$ & $\mathbf{z} = (\mathbf{x} - \mathbf{o}) \times 0.5$ & Shifted Penalized Function 1\\
                $f_{16}$\tnote{+} & $\begin{aligned}\frac{\pi}{n}[10\sin^2(\pi y_1)+\sum\nolimits_{i=1}^{n-1}(y_i-1)^2(1+10\sin^2(\pi y_{i+1}))+\\(y_n-1)^2]+\sum\nolimits_{i=1}^n u(z_i,10,100,4), y_i = 1+\frac{1}{4}(z_i+1)\end{aligned}$ & $\mathbf{z} = (\mathbf{x} - \mathbf{o}) \times 0.5$ & Shifted Penalized Function 2\\
                \hline			
            \end{tabular}
            \begin{tablenotes}\footnotesize
                \item [*] $\boldsymbol{o}$ is a shifting vector and $\boldsymbol{M}$ is a transformation matrix. $\boldsymbol{o}$ and $\boldsymbol{M}$ can be obtained from \cite{LiangQuSuganthanHernandez-Diaz2013ProblemDefinitionsand}.
                \item [+] $u(x,a,k,m)=\begin{cases}k(x-a)^m & \mathrm{for\:}x>a\\0& \mathrm{for\:}-a\leq x\leq a\\k(-x-a)^m & \mathrm{for\:}x<-a \end{cases}$.
            \end{tablenotes}
        \end{threeparttable}
	\end{center}
\end{table*}

\subsection{Other Modifications}

In order to fully exploit the utility of our proposed inter-molecular collision scheme, we also make some other modifications to the implementation of the CRO framework. 

In the canonical CRO design, the algorithm does not impose any mandatory constraints on the elementary reaction selection until the size of the population is reduced to one. The selection is controlled by the optimization parameters. When the population size is one, neither \textit{Inter} nor \textit{Syn} would occur as they need at least two molecules as inputs. However this behavior may potentially reduce the occurrence of \textit{Inter} and in return bias the ratio of \textit{Onwall} and \textit{Inter}. So we add a mandatory constraint to maintain this ratio.

In CRO/AC, the elementary reaction selection scheme works identically to the canonical design of CRO in the first step, i.e. a random number is first generated and compared with $collRate$. If the random number is larger, either \textit{Onwall} or \textit{Dec} will occur. Otherwise, an \textit{Inter} or \textit{Syn} will take place and here we impose the second mandatory condition. If the current population size is no larger than two, all later selection steps are skipped and an \textit{Inter} will occur. This modification can guarantee that the size of population is always larger than one, and there are always enough molecules in the container to perform an \textit{Inter} elementary reaction.

\section{Experimental Setting and Simulation Results} \label{sec:expsim}

In this section we will first introduce the benchmark functions adopted for performance evaluation and the experimental settings used. The simulation results, comparisons and discussions will also be presented.

\subsection{Benchmark Functions}

In order to evaluate our proposed CRO/AC, we conduct a series of simulations on 16 different benchmark functions. These benchmark functions are selected form the benchmark set proposed by Yao \textit{et al.} \cite{YaoLiuLin1999EvolutionaryProgrammingMade} and the latest benchmark problem set for the competition on real-parameter single objective optimization at CEC 2013 \cite{LiangQuSuganthanHernandez-Diaz2013ProblemDefinitionsand}. The former benchmark set has been adopted for testing performance by a wide range of metaheuristics in recent years \cite{HeWuSaunders2009GroupSearchOptimizer:}, and was adopted to test the performance of RCCRO \cite{LamLiYu2012RealCodedChemical}. As revealed in \cite{QinHuangSuganthan2009DifferentialEvolutionAlgorithm}, the current benchmark functions often suffer from two major problems: the global optimal points are located at the center of the search space, and the inter-dimensional correlation is weak. So we make a comprehensive test suite to resolve these problems by shifting and rotating some of the benchmark functions. The benchmark functions are listed in Table \ref{tbl:benchmark}. All benchmark functions are optimized in 30 dimensions, and the search spaces are defined to be $[-100, 100]$. The global optimum values of these benchmark functions are zero, and all simulation results smaller than $10^{-8}$ are considered to be zero \cite{LiangQuSuganthanHernandez-Diaz2013ProblemDefinitionsand}.

To evaluate the performance improvement of our proposed CRO/AC over CRO, we compare the simulation results between these two designs. Lam \textit{et al.} proposed three canonical CRO variants in \cite{LamLiYu2012RealCodedChemical}, where the major differences among them are the constraint handling schemes and synthesis operators. We use CRO/BP, CRO/HP, and CRO/BB to refer to RCCRO1, RCCRO2, and RCCRO3 described in \cite{LamLiYu2012RealCodedChemical}. We will compare the performance of CRO/AC with these three algorithms.

CRO/AC and all the CRO variants are implemented in Python 2.7 on Microsoft Windows 7. All simulations are performed on a computer with an Intel Core i7-3770 @ 3.4GHz CPU. In order to reduce statistical errors and generate statistically significant results, each benchmark function is repeated for 51 independent runs for each algorithm according to the suggestion by \cite{LiangQuSuganthanHernandez-Diaz2013ProblemDefinitionsand}. We use the maximum number of function evaluation (maxFE) as the termination criteria and maxFE is set to 300 000 for all benchmark functions, satisfying the requirement of \cite{LiangQuSuganthanHernandez-Diaz2013ProblemDefinitionsand}.

The parameters for all algorithms are set according to the recommendation of \cite{LamLiYu2012RealCodedChemical} for solving multimodal problems, i.e. population size is 20, \textit{stepsize} is 1, initial energy buffer is $10^5$, initial kinetic energy for molecules is $10^7$, molecular collision rate is 0.2, kinetic energy loss rate is 0.1, decomposition threshold is $1.5\times10^5$, and synthesis threshold is 10.

\subsection{Comparison of CRO/AC and Canonical CRO Variants}

\begin{table}
	\caption{Simulation Results of CRO/AC and Canonical CRO Variants}
	\label{tbl:croacres}
	\begin{center}
	    \scriptsize
        \begin{threeparttable}
            \begin{tabular}{ll|llll}
                \hline
                \multicolumn{2}{c|}{Function} & CRO/AC & CRO/BP & CRO/BB & CRO/HP \\
                \hline
\multirow{3}{*}{$f_{1}$} & Mean & \textbf{4.2111e-07} & 2.7374e-06 & 2.6009e-06 & 2.6684e-06 \\
 & Std. Dev. & 3.4578e-07 & 1.5487e-06 & 1.2265e-06 & 1.4386e-06 \\
 & T-test & - & \textit{-10.3216} & \textit{-12.0957} & \textit{-10.7402} \\ \hline
\multirow{3}{*}{$f_{2}$} & Mean & \textbf{8.0285e-06} & 1.8280e-05 & 1.8690e-05 & 1.9248e-05 \\
 & Std. Dev. & 7.1100e-06 & 8.4089e-06 & 9.3814e-06 & 9.9651e-06 \\
 & T-test & - & \textit{-6.5827} & \textit{-6.4043} & \textit{-6.4804} \\ \hline
\multirow{3}{*}{$f_{3}$} & Mean & \textbf{1.2540e-04} & 5.2846e-04 & 6.2956e-04 & 5.9767e-04 \\
 & Std. Dev. & 5.1818e-05 & 1.2148e-04 & 1.6374e-04 & 1.3829e-04 \\
 & T-test & - & \textit{-21.5802} & \textit{-20.7578} & \textit{-22.6129} \\ \hline
\multirow{3}{*}{$f_{4}$} & Mean & \textbf{3.9728e+01} & 1.2944e+07 & 4.6210e+06 & 6.8512e+06 \\
 & Std. Dev. & 1.9891e+01 & 7.0206e+07 & 3.0236e+07 & 2.3599e+07 \\
 & T-test & - & -1.3037 & -1.0807 & \textit{-2.0528} \\ \hline
\multirow{3}{*}{$f_{5}$} & Mean & \textbf{7.5571e+00} & 1.1503e+01 & 1.1177e+01 & 1.1305e+01 \\
 & Std. Dev. & 6.6117e-01 & 5.5733e-01 & 6.6192e-01 & 6.6029e-01 \\
 & T-test & - & \textit{-32.2656} & \textit{-27.3574} & \textit{-28.3605} \\ \hline
\multirow{3}{*}{$f_{6}$} & Mean & \textbf{7.8682e+00} & 1.1388e+01 & 1.1231e+01 & 1.1562e+01 \\
 & Std. Dev. & 8.0475e-01 & 7.0835e-01 & 6.6992e-01 & 5.5514e-01 \\
 & T-test & - & \textit{-23.2181} & \textit{-22.7117} & \textit{-26.7181} \\ \hline
\multirow{3}{*}{$f_{7}$} & Mean & \textbf{0.0000e+00} & 5.8231e+03 & 6.1206e+03 & 6.0672e+03 \\
 & Std. Dev. & 0.0000e+00 & 6.1178e+02 & 7.8931e+02 & 6.6152e+02 \\
 & T-test & - & \textit{-67.3051} & \textit{-54.8319} & \textit{-64.8529} \\ \hline
\multirow{3}{*}{$f_{8}$} & Mean & \textbf{0.0000e+00} & 5.7527e+03 & 5.6308e+03 & 5.5973e+03 \\
 & Std. Dev. & 0.0000e+00 & 6.7561e+02 & 1.1770e+03 & 6.4795e+02 \\
 & T-test & - & \textit{-60.2089} & \textit{-33.8292} & \textit{-61.0824} \\ \hline
\multirow{3}{*}{$f_{9}$} & Mean & \textbf{1.3400e+02} & 4.2171e+02 & 3.8941e+02 & 4.3384e+02 \\
 & Std. Dev. & 2.9142e+01 & 7.8572e+01 & 8.4037e+01 & 8.4771e+01 \\
 & T-test & - & \textit{-24.2770} & \textit{-20.3045} & \textit{-23.6523} \\ \hline
\multirow{3}{*}{$f_{10}$} & Mean & \textbf{1.4383e+02} & 4.3566e+02 & 4.1632e+02 & 4.5059e+02 \\
 & Std. Dev. & 3.9544e+01 & 8.9085e+01 & 8.8294e+01 & 9.0004e+01 \\
 & T-test & - & \textit{-21.1713} & \textit{-19.9162} & \textit{-22.0647} \\ \hline
\multirow{3}{*}{$f_{11}$} & Mean & 2.7991e+00 & 2.3064e+00 & 2.6168e+00 & \textbf{2.1902e+00} \\
 & Std. Dev. & 3.4417e+00 & 2.8145e+00 & 2.6376e+00 & 2.4651e+00 \\
 & T-test & - & 0.7837 & 0.2972 & 1.0171 \\ \hline
\multirow{3}{*}{$f_{12}$} & Mean & 1.0817e-02 & 4.7648e-03 & 4.0542e-03 & \textbf{1.9807e-03} \\
 & Std. Dev. & 1.2490e-02 & 1.2358e-02 & 6.6509e-03 & 4.3693e-03 \\
 & T-test & - & \textit{2.4357} & \textit{3.3796} & \textit{4.7222} \\ \hline
\multirow{3}{*}{$f_{13}$} & Mean & \textbf{2.8092e+00} & 1.9428e+01 & 1.0251e+01 & 1.2036e+01 \\
 & Std. Dev. & 2.8073e+00 & 2.4661e+01 & 1.1397e+01 & 1.3504e+01 \\
 & T-test & - & \textit{-4.7344} & \textit{-4.4832} & \textit{-4.7302} \\ \hline
\multirow{3}{*}{$f_{14}$} & Mean & \textbf{5.2333e+00} & 3.7536e+01 & 6.9319e+01 & 3.8191e+01 \\
 & Std. Dev. & 5.6115e+00 & 4.2015e+01 & 4.3951e+01 & 3.8917e+01 \\
 & T-test & - & \textit{-5.3885} & \textit{-10.2274} & \textit{-5.9271} \\ \hline
\multirow{3}{*}{$f_{15}$} & Mean & \textbf{5.7647e-08} & 2.6110e-07 & 2.1121e-07 & 2.3093e-07 \\
 & Std. Dev. & 1.7199e-07 & 2.0579e-07 & 1.9243e-07 & 1.8260e-07 \\
 & T-test & - & \textit{-5.3640} & \textit{-4.2072} & \textit{-4.8848} \\ \hline
\multirow{3}{*}{$f_{16}$} & Mean & \textbf{7.4478e+00} & 1.6633e+01 & 6.9264e+00 & 1.5481e+01 \\
 & Std. Dev. & 6.9020e+00 & 1.1806e+01 & 7.3114e+00 & 9.8198e+00 \\
 & T-test & - & \textit{-4.7492} & 0.3667 & \textit{-4.7324} \\ \hline
            \end{tabular}
        \end{threeparttable}
	\end{center}
\end{table}

The mean values, standard deviations, and Student's t-test results obtained by CRO/AC and the three canonical CRO variants are presented in Table \ref{tbl:croacres}. The mean values in bold indicate superiority of the corresponding algorithm over the others. The t-test results are the calculated t-statistic, where a negative value means that CRO/AC outperforms the corresponding algorithm. The t-test values in italic indicate that the advantage is significant at a confidence level of 95\%. From the simulation results the following key points can be observed:

\begin{itemize}
\item CRO/AC performs better than all canonical CRO variants in most of the tested benchmark functions when comparing the mean simulation results. CRO/AC outperforms the others in 14 functions, and its performance in the remaining two functions ($f_{11}$ and $f_{12}$) are comparable to the CRO variants.

\item The t-test results further support our previous observation. CRO/AC outperforms other algorithms in all but three functions ($f_4$, $f_{11}$ and $f_{12}$). In $f_4$, CRO/AC outperforms CRO/HP and performs similarly with CRO/BP and CRO/BB. In the meantime CRO/AC generates more stable simulation results. In $f_{11}$, CRO/AC generates a slightly worse mean result, but the t-test results indicate no significant advantage of the other algorithms.
\end{itemize}

In terms of the computational time, a very small amount of additional time is required for our inter-molecular adaptive collision scheme. The total extra time is around 2\%--7\% of the computational time needed by the canonical CRO variants. As the benchmark functions we adopt are relatively less computationally intensive compared with most real-world problems, we believe that the extra time used by CRO/AC is not critical considering the significant performance improvement.

\subsection{Analysis on the Two-step Inter-molecular Ineffective Collision Operator and Adaptive Collision Scheme}

\begin{table}
	\caption{Simulation Results of CRO/AC variants}
	\label{tbl:croacvar}
	\begin{center}
	    \scriptsize
        \begin{threeparttable}
            \begin{tabular}{ll|lll}
                \hline
                \multicolumn{2}{c|}{Function} & CRO/AC & CRO/AC/0.2 & CRO/AC/1step \\
                \hline
\multirow{3}{*}{$f_{1}$} & Mean & 4.2111e-07 & \textbf{8.9465e-08} & 3.8261e-07 \\
 & Std. Dev. & 3.4578e-07 & 6.9765e-08 & 2.7951e-07 \\
 & T-test & - & \textit{6.6481} & 0.6123 \\ \hline
\multirow{3}{*}{$f_{2}$} & Mean & 8.0285e-06 & \textbf{2.1730e-06} & 1.3225e-05 \\
 & Std. Dev. & 7.1100e-06 & 2.1997e-06 & 7.9482e-06 \\
 & T-test & - & \textit{5.5633} & \textit{-3.4455} \\ \hline
\multirow{3}{*}{$f_{3}$} & Mean & 1.2540e-04 & \textbf{6.0034e-05} & 1.9051e-04 \\
 & Std. Dev. & 5.1818e-05 & 2.3344e-05 & 5.8835e-05 \\
 & T-test & - & \textit{8.1321} & \textit{-5.8727} \\ \hline
\multirow{3}{*}{$f_{4}$} & Mean & \textbf{3.9728e+01} & 2.3252e+04 & 9.4010e+01 \\
 & Std. Dev. & 1.9891e+01 & 1.1502e+05 & 4.1884e+01 \\
 & T-test & - & -1.4270 & \textit{-8.2781} \\ \hline
\multirow{3}{*}{$f_{5}$} & Mean & \textbf{7.5571e+00} & 9.1927e+00 & 8.3137e+00 \\
 & Std. Dev. & 6.6117e-01 & 8.0608e-01 & 6.7715e-01 \\
 & T-test & - & \textit{-11.0930} & \textit{-5.6529} \\ \hline
\multirow{3}{*}{$f_{6}$} & Mean & \textbf{7.8682e+00} & 9.4500e+00 & 8.5155e+00 \\
 & Std. Dev. & 8.0475e-01 & 7.2286e-01 & 6.8856e-01 \\
 & T-test & - & \textit{-10.3400} & \textit{-4.3216} \\ \hline
\multirow{3}{*}{$f_{7}$} & Mean & \textbf{0.0000e+00} & 6.7445e+02 & 6.6541e+03 \\
 & Std. Dev. & 0.0000e+00 & 1.3551e+03 & 7.5036e+02 \\
 & T-test & - & \textit{-3.5193} & \textit{-62.7052} \\ \hline
\multirow{3}{*}{$f_{8}$} & Mean & \textbf{0.0000e+00} & 3.7914e+02 & 6.6298e+03 \\
 & Std. Dev. & 0.0000e+00 & 1.0612e+03 & 8.1775e+02 \\
 & T-test & - & \textit{-2.5264} & \textit{-57.3279} \\ \hline
\multirow{3}{*}{$f_{9}$} & Mean & \textbf{1.3400e+02} & 2.1140e+02 & 1.9889e+02 \\
 & Std. Dev. & 2.9142e+01 & 4.7866e+01 & 3.8175e+01 \\
 & T-test & - & \textit{-9.7669} & \textit{-9.5546} \\ \hline
\multirow{3}{*}{$f_{10}$} & Mean & \textbf{1.4383e+02} & 2.0337e+02 & 2.1631e+02 \\
 & Std. Dev. & 3.9544e+01 & 5.4420e+01 & 4.2253e+01 \\
 & T-test & - & \textit{-6.2583} & \textit{-8.8565} \\ \hline
\multirow{3}{*}{$f_{11}$} & Mean & 2.7991e+00 & \textbf{4.4066e-01} & 7.3760e+00 \\
 & Std. Dev. & 3.4417e+00 & 8.6072e-01 & 5.0705e+00 \\
 & T-test & - & \textit{4.7007} & \textit{-5.2810} \\ \hline
\multirow{3}{*}{$f_{12}$} & Mean & 1.0817e-02 & \textbf{1.0379e-02} & 1.1223e-02 \\
 & Std. Dev. & 1.2490e-02 & 1.3075e-02 & 1.6374e-02 \\
 & T-test & - & 0.1712 & -0.1393 \\ \hline
\multirow{3}{*}{$f_{13}$} & Mean & \textbf{2.8092e+00} & 6.3203e+00 & 6.6642e+00 \\
 & Std. Dev. & 2.8073e+00 & 8.2816e+00 & 9.4389e+00 \\
 & T-test & - & \textit{-2.8391} & \textit{-2.7681} \\ \hline
\multirow{3}{*}{$f_{14}$} & Mean & \textbf{5.2333e+00} & 1.9807e+01 & 7.4475e+00 \\
 & Std. Dev. & 5.6115e+00 & 2.2535e+01 & 1.1233e+01 \\
 & T-test & - & \textit{-4.4374} & -1.2469 \\ \hline
\multirow{3}{*}{$f_{15}$} & Mean & 5.7647e-08 & \textbf{1.1126e-08} & 9.2002e-08 \\
 & Std. Dev. & 1.7199e-07 & 3.2826e-08 & 1.6632e-07 \\
 & T-test & - & 1.8787 & -1.0153 \\ \hline
\multirow{3}{*}{$f_{16}$} & Mean & \textbf{7.4478e+00} & 1.2715e+01 & 1.7369e+01 \\
 & Std. Dev. & 6.9020e+00 & 1.0369e+01 & 1.4167e+01 \\
 & T-test & - & \textit{-2.9904} & \textit{-4.4514} \\ \hline
            \end{tabular}
        \end{threeparttable}
	\end{center}
\end{table}

Section \ref{sec:CROAC} introduced our proposed inter-molecular ineffective collision operator, which is a two-step manipulation method. In this section we will show the necessity of incorporating the second step, i.e. (\ref{eqn:interstep2}). We construct a CRO/AC variant with one-step inter-molecular ineffective collision operator (CRO/AC/1step) which removes the second step of manipulating the input molecules. We also proposed an adaptive collision scheme which adaptively changes the value for the molecular collision rate. We construct CRO/AC/0.2 which sets the collision rate constantly to 0.2 during the entire search process. The same experimental settings and benchmark functions are applied to these algorithms, and the simulation results are presented in Table \ref{tbl:croacvar}. From these test results, we have the following observations:

\begin{itemize}
\item Both the two-step manipulation method and the adaptive collision scheme contribute to the performance improvement of CRO/AC. CRO/AC generates 10 best mean values among the 16 benchmark functions while CRO/AC/0.2 produces the best mean values pf the other six.

\item CRO/AC/0.2 tends to generate outstanding results in the uni-modal optimization functions ($f_1$--$f_4$), while CRO/AC is generally superior at solving multi-modal functions ($f_5$--$f_{16}$)

\item The less competitive results generated by CRO/AC/1step are potentially due to pre-mature convergence, which is caused by missing the second step manipulation that allows molecules to jump out of local optima.
\end{itemize}

\begin{figure}
    \includegraphics[width=0.48\textwidth]{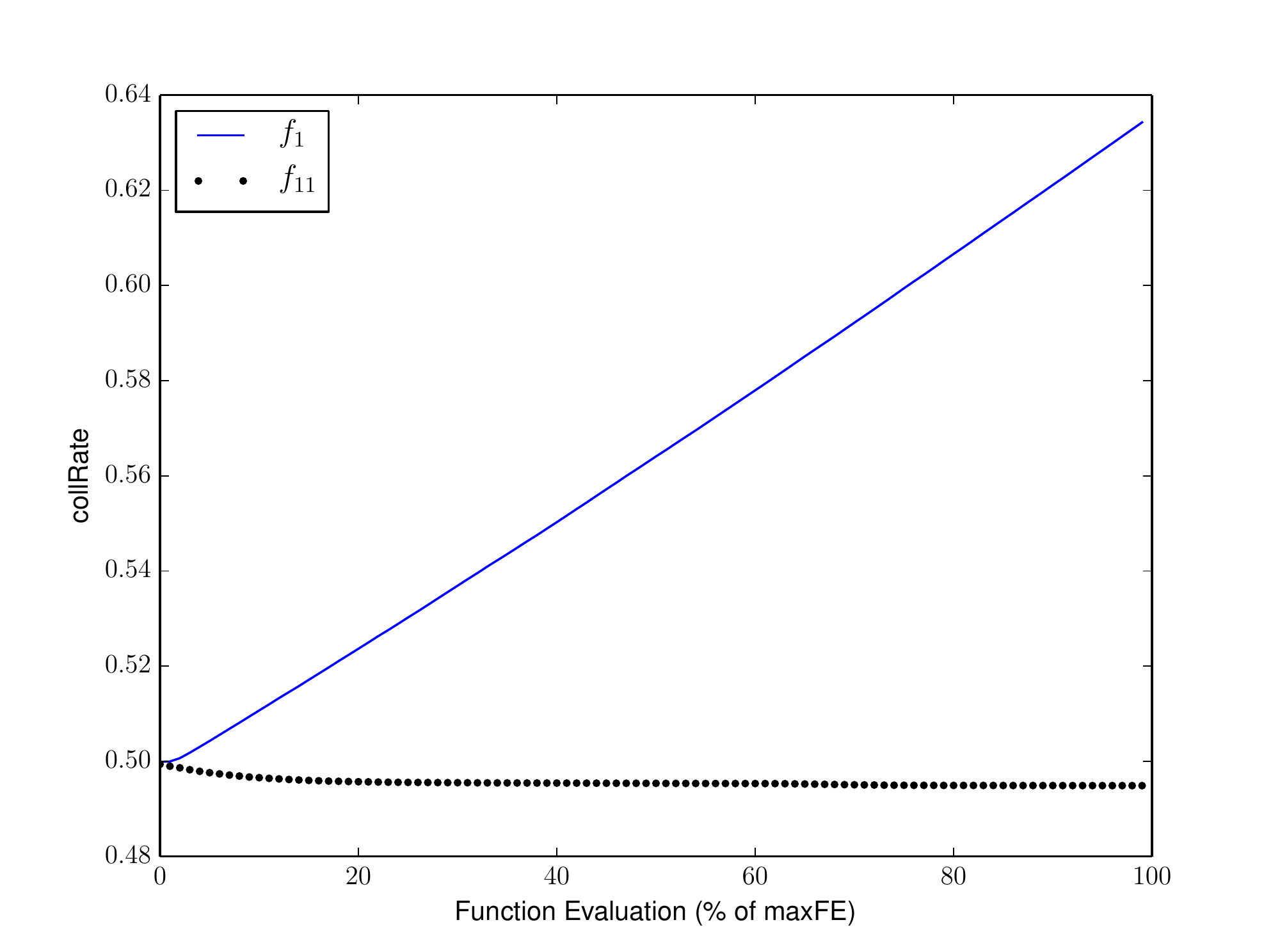}
    \caption{$collRate$ Changes of $f_1$ and $f_{11}$}
    \label{fig:collratechange}
\end{figure}

The effect of the adaptive collision scheme can also be revealed by Fig. \ref{fig:collratechange}, where the changes on $collRate$ of $f_1$ and $f_5$ are plotted. From the figure we can see the adaptive scheme can adjust $collRate$ to different values according to the feedback from the optimization process. 

\section{Conclusion} \label{sec:conclusion}

CRO is a recently proposed simple and powerful metaheuristic optimization method which mimics the interactions and transformations of molecules in chemical reactions to search for the global optimum. CRO has been applied to solve both combinatorial and continuous optimization problems. In the canonical design of CRO for solving continuous optimization problems, the functionalities of on-wall ineffective collision and inter-molecular ineffective collision overlap, and this potentially hampers the performance of CRO. In this paper we propose a novel inter-molecular operator to overcome this drawback. This operator changes the values on multiple dimensions of a solution at one time, and is more efficient at solving problems with strong inter-dimensional correlation than the canonical design. In order to fully utilize the different optimization features of this new operator, we also devise an adaptive collision scheme to control the ratio of different elementary reactions. This scheme learns from the optimization feedback and adjusts the $collRate$ parameter of CRO to adapt the algorithm to different problems. We incorporate the new operator and the adaptive scheme into the CRO framework and propose a new algorithm CRO/AC.

To examine the performance improvement of CRO/AC over canonical CRO, we perform a series of simulations on a wide range of different benchmark functions. The simulation results indicate the superiority of our proposed CRO/AC over all the compared CRO variants. We also analyze the performance improvement contribution made by the new operator and the adaptive scheme, and find both of them are essential for CRO/AC to achieve better performance than the canonical CRO.

In the future we will perform a systematic analysis on the parameter selection for CRO/AC. All simulation in this paper adopted the parameter settings devised in \cite{LamLiYu2012RealCodedChemical}, which was originally designed for the canonical CRO and may be not suitable for CRO/AC. It is also an interesting topic to combine CRO/AC with the $stepSize$ adaptation scheme proposed in \cite{LamLiYu2012RealCodedChemical}. Last but not least, we will apply CRO/AC to real-world practical optimization problems to perform an overall study of the performance of CRO/AC.

\section*{Acknowledgement}
This research is supported in part by the University of Hong Kong Strategic Research Theme on Computation and Information. A.Y.S. Lam is supported in part by the Faculty Research Grant of Hong Kong Baptist University, under Grant No. FRG2/13-14/045.

\bibliographystyle{IEEEtran}
\bibliography{IEEEabrv,../../../bib/publications}

\begin{thebibliography}{10}
\providecommand{\url}[1]{#1}
\csname url@samestyle\endcsname
\providecommand{\newblock}{\relax}
\providecommand{\bibinfo}[2]{#2}
\providecommand{\BIBentrySTDinterwordspacing}{\spaceskip=0pt\relax}
\providecommand{\BIBentryALTinterwordstretchfactor}{4}
\providecommand{\BIBentryALTinterwordspacing}{\spaceskip=\fontdimen2\font plus
\BIBentryALTinterwordstretchfactor\fontdimen3\font minus
  \fontdimen4\font\relax}
\providecommand{\BIBforeignlanguage}[2]{{%
\expandafter\ifx\csname l@#1\endcsname\relax
\typeout{** WARNING: IEEEtran.bst: No hyphenation pattern has been}%
\typeout{** loaded for the language `#1'. Using the pattern for}%
\typeout{** the default language instead.}%
\else
\language=\csname l@#1\endcsname
\fi
#2}}
\providecommand{\BIBdecl}{\relax}
\BIBdecl

\bibitem{LamLi2010ChemicalReactionInspired}
A.~Y.~S. Lam and V.~O.~K. Li, ``Chemical-reaction-inspired metaheuristic for
  optimization,'' \emph{{IEEE} Trans. Evol. Comput.}, vol.~14, no.~3, pp.
  381--399, 2010.

\bibitem{LamLi2012ChemicalReactionOptimization:}
------, ``Chemical reaction optimization: A tutorial,'' \emph{Memetic
  Computing}, vol.~4, no.~1, pp. 3--17, 2012.

\bibitem{YuLiLam2012SensorDeploymentAir}
J.~J.~Q. Yu, V.~O.~K. Li, and A.~Y.~S. Lam, ``Sensor deployment for air
  pollution monitoring using public transportation system,'' in \emph{Proc.
  IEEE Congress on Evolutionary Computation (CEC)}, Brisbane, Australia, Jun.
  2012, pp. 1--7.

\bibitem{YuLiLam2013OptimalV2GScheduling}
------, ``Optimal {V2G} scheduling of electric vehicles and unit commitment
  using chemical reaction optimization,'' in \emph{Proc. IEEE Congress on
  Evolutionary Computation (CEC)}, Cancun, Mexico, Jun. 2013, pp. 392--399.

\bibitem{LamLiYu2012RealCodedChemical}
A.~Y.~S. Lam, V.~O.~K. Li, and J.~J.~Q. Yu, ``Real-coded chemical reaction
  optimization,'' \emph{{IEEE} Trans. Evol. Comput.}, vol.~16, no.~3, pp.
  339--353, 2012.

\bibitem{YuLamLi2011EvolutionaryArtificialNeural}
J.~J.~Q. Yu, A.~Y.~S. Lam, and V.~O.~K. Li, ``Evolutionary artificial neural
  network based on chemical reaction optimization,'' in \emph{Proc. IEEE
  Congress on Evolutionary Computation (CEC)}, New Orleans, LA, U.S., Jun.
  2011, pp. 2083--2090.

\bibitem{SunLamLiXuYu2012ChemicalReactionOptimization}
Y.~Sun, A.~Y.~S. Lam, V.~O.~K. Li, J.~Xu, and J.~J.~Q. Yu, ``Chemical reaction
  optimization for the optimal power flow problem,'' in \emph{Proc. IEEE
  Congress on Evolutionary Computation (CEC)}, Brisbane, Australia, Jun. 2012,
  pp. 1--8.

\bibitem{LamLiYu2013PowerControlledCognitive}
A.~Y.~S. Lam, V.~O.~K. Li, and J.~J.~Q. Yu, ``Power-controlled cognitive radio
  spectrum allocation with chemical reaction optimization,'' \emph{{IEEE}
  Trans. Wireless Commun.}, vol.~12, no.~7, pp. 3180--3190, 2013.

\bibitem{YuLamLi2012RealcodedChemical}
J.~J.~Q. Yu, A.~Y.~S. Lam, and V.~O.~K. Li, ``Real-coded chemical reaction
  optimization with different perturbation functions,'' in \emph{Proc. IEEE
  Congress on Evolutionary Computation (CEC)}, Brisbane, Australia, Jun. 2012,
  pp. 1--8.

\bibitem{QinHuangSuganthan2009DifferentialEvolutionAlgorithm}
A.~Qin, V.~Huang, and P.~Suganthan, ``Differential evolution algorithm with
  strategy adaptation for global numerical optimization,'' \emph{{IEEE} Trans.
  Evol. Comput.}, vol.~13, no.~2, pp. 398--417, 2009.

\bibitem{XuLamLi2011ChemicalReactionOptimization}
J.~Xu, A.~Y.~S. Lam, and V.~O.~K. Li, ``Chemical reaction optimization for task
  scheduling in grid computing,'' \emph{{IEEE} Trans. Parallel Distrib. Syst.},
  vol.~22, no.~10, pp. 1624--1631, Jan. 2011.

\bibitem{BeyerDeb2001SelfAdaptiveFeatures}
H.-G. Beyer and K.~Deb, ``On self-adaptive features in real-parameter
  evolutionary algorithms,'' \emph{{IEEE} Trans. Evol. Comput.}, vol.~5, no.~3,
  pp. 250--270, Jun. 2001.

\bibitem{LiangQuSuganthanHernandez-Diaz2013ProblemDefinitionsand}
J.~J. Liang, B.-Y. Qu, P.~N. Suganthan, and A.~G. Hern¨¢ndez-D¨ªaz, ``Problem
  definitions and evaluation criteria for the {CEC} 2013 special session and
  competition on real-parameter optimization,'' Computational Intelligence
  Laboratory, Zhengzhou University, Zhengzhou China and Nanyang Technological
  University, Singapore, Technical Report 201212, 2013.

\bibitem{YaoLiuLin1999EvolutionaryProgrammingMade}
X.~Yao, Y.~Liu, and G.~Lin, ``Evolutionary programming made faster,''
  \emph{{IEEE} Trans. Evol. Comput.}, vol.~3, no.~2, pp. 82--102, Aug. 1999.

\bibitem{HeWuSaunders2009GroupSearchOptimizer:}
S.~He, Q.~H. Wu, and J.~R. Saunders, ``Group search optimizer: An optimization
  algorithm inspired by animal searching behavior,'' \emph{{IEEE} Trans. Evol.
  Comput.}, vol.~13, no.~5, pp. 973--990, Aug. 2009.

\end{thebibliography}

\end{document}